\documentclass[letterpaper, 10 pt, conference]{ieeeconf}  
\IEEEoverridecommandlockouts                              
\usepackage{amssymb,amsmath}

\usepackage{amsmath}
\usepackage[american]{babel}
\usepackage{graphicx}
\usepackage[utf8x]{inputenc}
\usepackage{dsfont}
\usepackage{caption}
\usepackage{subcaption}
\usepackage{todonotes}
\usepackage{xspace}
\usepackage{url}
\usepackage{wrapfig}
\usepackage{algpseudocode}
\usepackage{gensymb}
\usepackage[pdfborder={0 0 0}]{hyperref} 
\usepackage{siunitx}
\usepackage{color}
\usepackage{flushend}

\DeclareCaptionType{copyrightbox}

\title{\LARGE \bf Distributed Cohesive Control for Robot Swarms:\\
Maintaining Good Connectivity in the Presence of Exterior Forces}

\author{
Dominik Krupke$^{1}$,
Maximilian Ernestus$^{1}$, 
Michael Hemmer$^{1}$, and 
S\'andor~P.~Fekete$^{1}$%
\thanks{$^{1}$S. Fekete, M. Hemmer, and D. Krupke are with the Computer Science Department, TU Braunschweig, Braunschweig, Germany, which is also where
M. Ernestus carried out his work. {\tt\small maximilian@ernestus.de, s.fekete@tu-bs.de, mhsaar@gmail.com, d.krupke@tu-bs.de}}%
}

\begin{document}

\graphicspath{{./fig/}}

\newcommand{\bigO}{\operatorname{O}}
\newcommand{\eg}{e.\,g.\xspace}
\newcommand{\ie}{i.\,e.\xspace}
\newcommand{\NP}{\textit{NP}}
\newcommand{\Obda}{W.\,l.\,o.\,g.\xspace}
\newcommand{\obda}{w.\,l.\,o.\,g.\xspace} 
\newcommand{\R}{\mathds{R}}
\newcommand{\Z}{\mathds{Z}}

\newcommand{\email}[1]{\href{mailto:#1}{#1}}
\newcommand{\ignore}[1]{}

\newtheorem{algorithm}{Algorithm}
\newcommand{\localpheromone}[1][]{d_\text{#1}}
\newcommand{\emitterset}{\mathcal{E}}
\newcommand{\minemitterdist}{v}
\newcommand{\globalhopcount}{l}
\newcommand{\emitterdir}{\dot{\globalhopcount}}
\newcommand{\robotset}{\mathcal{R}}
\newcommand{\pheromonegraph}{\mathfrak{P}}
\newcommand{\localpheromonet}{\localpheromone^t}
\newcommand{\localpheromonetplus}{\localpheromone^{t+1}}
\newcommand{\globalhopcountt}{\globalhopcount^t}
\newcommand{\globalhopcounttplus}{\globalhopcount^{t+1}}
\newcommand{\leaderset}{\mathcal{L}}

\newcommand{\Base}{\textsc{Base}\xspace}
\newcommand{\Lead}{\textsc{Lead}\xspace}
\newcommand{\All}{\textsc{All}\xspace}

\maketitle
\thispagestyle{empty}
\pagestyle{empty}

\begin{abstract}
We present a number of powerful local mechanisms for maintaining
a dynamic swarm of robots with limited capabilities
and information, in the presence of external forces and permanent node failures. 
We propose a set of local {\em continuous} algorithms that together produce a
generalization of a 
Euclidean Steiner tree. At any stage, the resulting overall shape achieves 
a good compromise between local thickness, global connectivity, and flexibility to 
further continuous motion of the terminals. The resulting swarm behavior 
scales well, is robust against node failures, and performs
close to the best known approximation bound for a corresponding centralized
static optimization problem.
\end{abstract}



\section{Introduction}
\label{sec:introduction}

Consider a swarm of robots that needs to remain connected. 
There is no central control and no knowledge of the overall environment. 
This environment is hostile: The swarm is being pulled apart by external forces, 
stretching it into a number of different directions, so it is in danger of breaking up.
Individual robots are weak, with limited sensing, limited communication, and limited connectivity;
even worse, each robot's expected lifetime is limited by random, permanent failures, which may 
destroy connectedness and functioning of the swarm as a whole. 
How can we achieve coordinated dynamic swarm behavior without centralized coordination? 
How can we employ each robot as much as possible, without depending on it if it fails?
How can we balance overall flexibility and robustness to deal with the hostile environment?

\begin{figure}[t]
  \centering
  \includegraphics[width=0.9\columnwidth]{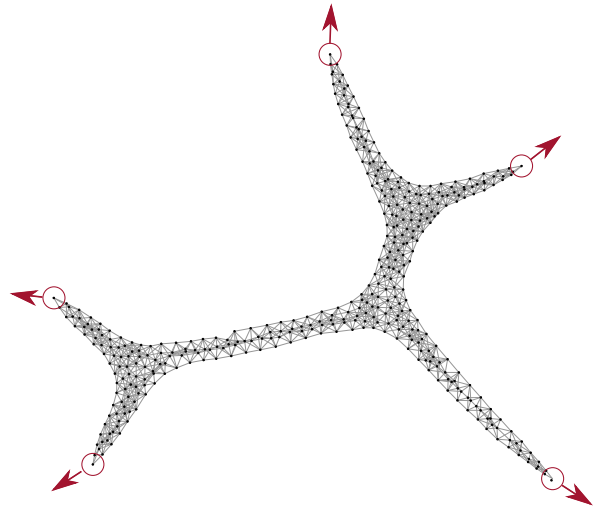}
  \caption{A robust robot swarm emulating a Steiner tree between five diverging attachment points.}
  \label{fig:experimentpicture}
\end{figure}

In this paper, we study swarm mechanisms that achieve these conflicting goals.
Just like in the paper by Lee and McLurkin~\cite{lee2014distributed}, we aim for algorithms that
(1) maintain connectivity, (2) are fully distributed, and
(3) achieve cohesiveness, i.e., a well-coordinated behavior and
state for all robots.
While \cite{lee2014distributed} present a
set of rules (based on crucial elements such as boundary recognition and boundary forces~\cite{McLurkin})
that achieve a ``fat'', well-rounded swarm shape even in the presence
of obstacles, this is no longer desirable in the presence of multiple outside forces
that pull the swarm apart, as illustrated in Figure~\ref{fig:experimentpicture}. As a consequence,
we formulate a new and additional goal: (4) achieve robust and adaptive overall swarm behavior,
even in the presence of external forces and node failures.

We present a combination of distributed boundary forces, density control and thickness regulation that go
beyond~\cite{lee2014distributed} by providing results for property (4). We achieve
a significant stability improvement over this and other previous approaches to flocking behavior, allowing
us to face scenarios for which even the corresponding centralized, static problems are NP-hard.
In a setting in which multiple dynamic terminals have to remain connected by
a generalized Steiner network with limited communication range, we achieve 
a performance that is comparable to the best worst-case guarantee of a theoretical, centralized approximation algorithm.

\subsection{Related Work.}
\label{ssec:related_work}
One of the earliest works on flocking is Reynold's pioneering work \cite{Reynolds}.
In recent years, a considerable number of aspects and objectives have extended this perspective.
We highlight only some of the ensuing papers, showing how they differ from our perspective.

A basic component of flocking is volumetric control, as presented by 
Spears~\cite{Spears}: robots use local potential field controllers (with attractive and repulsive
forces) for constructing a regular lattice with a corresponding base density~\cite{Olfati-Saber2006, andrew}.  
This does not necessarily preserve {\em connectivity}~\cite{Balch, heyes, Spears}. While the latter can be side-stepped
by simply assuming that robots are always connected~\cite{hong}, we aim for connectivity
as a requirement, which is vital in a fully distributed setting in which deterministic recovery from disconnectedness
may be impossible.

Some of the ideas of Olfati-Saber~\cite{Olfati-Saber2006} form the basis of our work and are discussed
in more detail further down. In~\cite{Olfati-Saber2006} and other work, however, robots do utilize gobal information, 
e.g., the position of a guide robot in a shared coordinate frame~\cite{Olfati-Saber2006, yao, hung, hung2} 
or environmental potential~\cite{Gazi}.
Instead of the potentials, Cortes~et.~al.~\cite{cortes} and Magnus~et.~al.~\cite{magnus} 
used Voronoi tessellation.
This is based on a density function, requiring global information for covering a region.
Overall, this differs from our objective of developing methods that are {\em fully
distributed}, aiming for collective mechanisms for complex group
behavior that go beyond relatively simple objectives~\cite{eric}, but also for
systems that are robust against partial hardware failures~\cite{kamimura}. 

The final property is ``cohesiveness'' of the overall swarm: all robots
should maintain a unified state, such as desired distance or
orientation; see~\cite{Olfati-Saber2006} for a formal definition. 
As described in~\cite{McLurkin}, detecting and maintaining a swarm boundary is of particular
importance for maintaining swarm cohesiveness and connectedness.
This is based on and related to work in the field of 
wireless sensor networks (WSNs), which has considered many geometric settings in which a 
large swarm of stationary nodes is faced with the task of achieving a large-scale overall goal, 
while the individual components can only operate locally, 
based on limited individual capabilities and information (\cite{Fekete2006}, \cite{Kroller2006}). 
In addition to the work on swarm robotics described above, there is a large body of theoretical 
work on geometric swarm behavior; for lack of space, we only mention
Chazelle~\cite{chaz} for flocking behavior, and Fekete et al.~\cite{Fekete2006,Kroller2006} for geometric
algorithms for static sensor networks, including distributed boundary detection.

Beyond the involved properties and paradigm, the overall goal for the swarm can also be described
as a distributed optimization problem: Maintain a generalized Steiner tree with limited edge lengths
that connects a moving set of terminals.
To the best of our knowledge, only Hamann and Wörn \cite{raey}
have explicitly considered the construction of Steiner trees by a robot swarm.
For static terminals, they start with an exploratory network; as soon as 
all terminals are connected, only best paths are kept and locally optimized.

Even in a centralized and static setting with full information, we have to deal with a generalization
of the well-known NP-hard problem of finding a good Steiner tree~\cite{garey1977complexity}.
More specifically, we are faced with the {\em relay placement problem}: the input is a set of 
sensors and a number $r \ge 1$, the communication range of a relay. 
The objective is to place a minimum number of relays so that between every pair
of sensors is connected by a path \emph{through sensors and/or relays}.
The best known theoretical performance bound for this NP-hard problem was given by Efrat et al.~\cite{efg-iaarp-08},
who presented a $3.11$-approximation algorithm; they also showed a worst-case lower bound of 3
for a large class of approximation algorithms.
For a fixed number of available relays, this turns into our problem of maximizing the achievable networks size,
with matching approximation factor.

More specific references are given in Section~\ref{sec:alg:base}, where they are used as
building blocks.

\subsection{Our contribution}
We propose a set of local, self-stabilizing algorithms that maintain a dynamic and robust network between leader robots.
The algorithms ensure that the swarm adopts the directions of multiple leaders, while preserving a uniform
thickness along the edges of the Steiner tree.
We demonstrate the usefulness of this approach by simulations with a swarm of 400 robots, five
leaders and various failure rates, by showing that the resulting performance is comparable to the
theoretical worst-case ratio.


\section{Preliminaries}
\label{sec:problem_definition}
We consider a finite set of robots $\mathcal{R}$. A subset
$\mathcal{L}\subsetneq \mathcal{R}, |\mathcal{L}|\ll |\mathcal{R}|$ of them
is forced to pursue externally controlled trajectories. For simplicity,
we call these {\em leader robots}; note that they have no control over their trajectories,
so they have no chance to keep the swarm coherent.
Instead, we want the remaining 
robots $\mathcal{R}\setminus \mathcal{L}$ to maintain a dynamic and robust network that keeps 
the swarm connected, even in the presence of random robot failures and arbitrary leader movements.
Thus, the overall shape of the swarm should form a ``thick'' Steiner 
tree among the leaders with the 
robots $\mathcal{R}\setminus \mathcal{L}$ evenly distributed along the edges, as shown 
in Figure~\ref{fig:experimentpicture}.


Robots have the shape of circles; two of them are connected 
when within a maximum distance and with an unobstructed line of sight.
Robots know the relative positions and orientations of their neighbors 
and can communicate asynchronously.
Each robot has a unique ID; leader IDs are easily made known to all others.
Robot's translations and rotations 
are limited in velocity and acceleration.
Communication is possible by broadcasting to immediate neighbors.

The perception of all robots is local; however, due to the known position and orientation 
difference, each robot can transform vectors of its neighbors to its own coordinate system.
We avoid multi-hop transformations to keep errors small; however, aggregate information is forwarded.



\section{Algorithm}
\label{sec:alg}
The proposed approach consists of a set of local self-stabilizing mechanisms that either detect a condition or induce a force.
The weighted sum of the induced forces determines the robot motion; input for the local mechanisms
of the local state and environment of the robot, output is a value for current
robot motion. In principle, these mechanisms are continuous. 
(Our simulator described later updates at 60 Hz.)



We first discuss the base behavior of the robots in Section~\ref{sec:alg:base}; because it has trouble with generating a non-convex swarm shape, it 
limits the flexibility of the swarm in the presence of external forces.
This is subsequently improved by leader forces, stability improvement and thickness contraction.

\subsection{Base Behavior}
\label{sec:alg:base}
\noindent
Our base behavior consists of three components: 
\begin{itemize}
\item[(i)] The \emph{flocking algorithm} of Olfati-Saber~\cite{Olfati-Saber2006}
considers regular distribution and movement consensus.  The
algorithm is a stateless equation based on potential fields and is proven to
converge. It uses three rules: Attraction to neighbors, repulsion
from too close neighbors, and adaption to the velocity of neighbors. 
We slightly modified the algorithm for better response to additional forces.

\item[(ii)] An extended version of the \emph{boundary detection} algorithm of McLurkin
and Demaine~\cite{McLurkin}, which determines if a robot lies
on the boundary and also identifies small holes by using the average angle.
In principle, the method allows the robots to distinguish exterior and
interior boundaries and determine their size, but the limited precision and the
convergence time limit this usage, so we only use it to detect and ignore small holes.
Doing the latter is crucial for thickness and density computation, see Section~\ref{sec:alg:stability}.

\item[(iii)] The \emph{boundary tension} of Lee and
McLurkin~\cite{lee2014distributed}, which straightens and minimizes the
boundary of the swarm. This is done by simply pushing boundary robots to the
middle of its two boundary neighbors.
\end{itemize}

\begin{figure}[tbh]
	\centering
	\begin{subfigure}[b]{0.2\textwidth}
	\includegraphics[width=\textwidth]{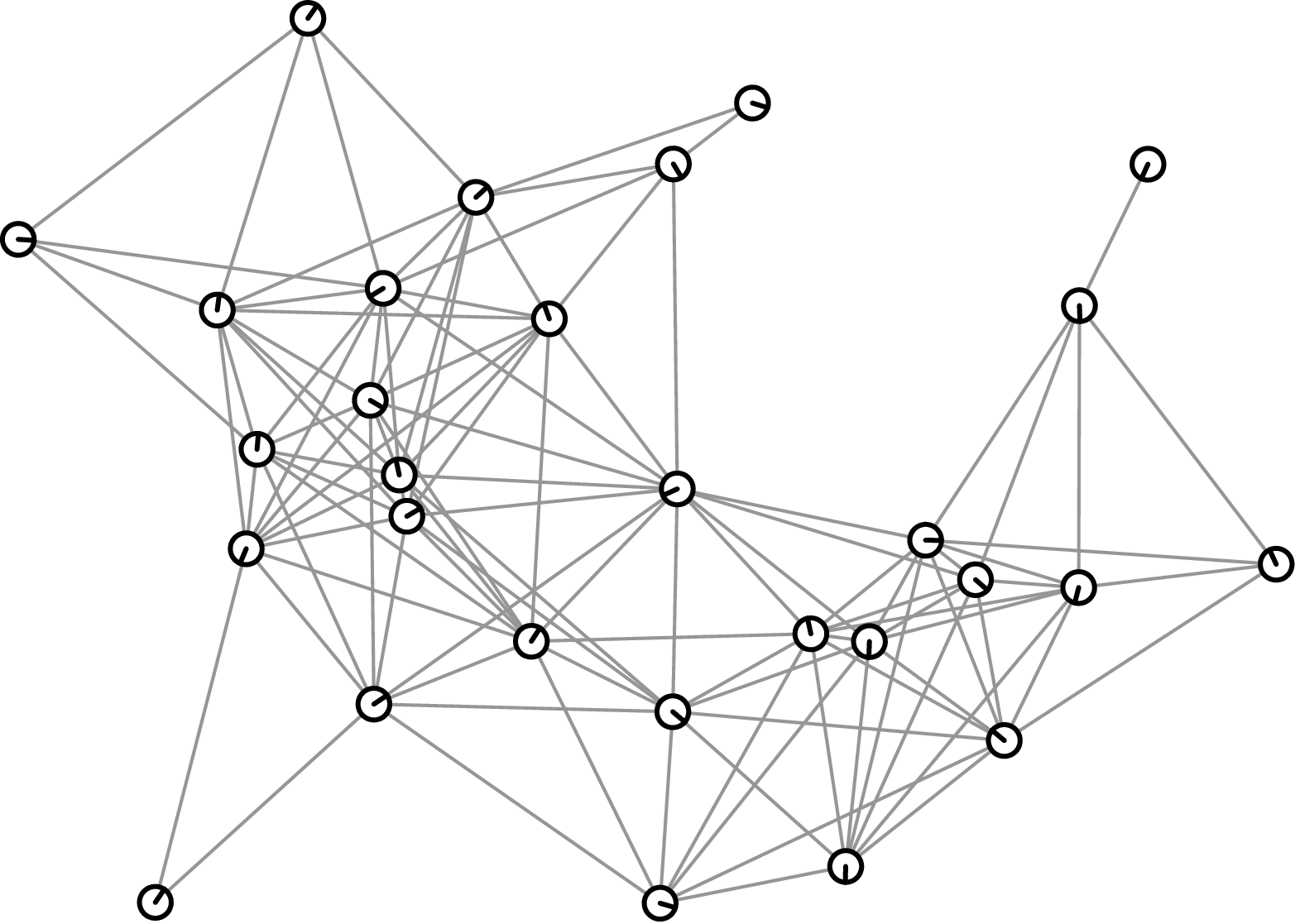}
	\caption{Before}
\end{subfigure}
\begin{subfigure}[b]{0.2\textwidth}
	\includegraphics[width=\textwidth]{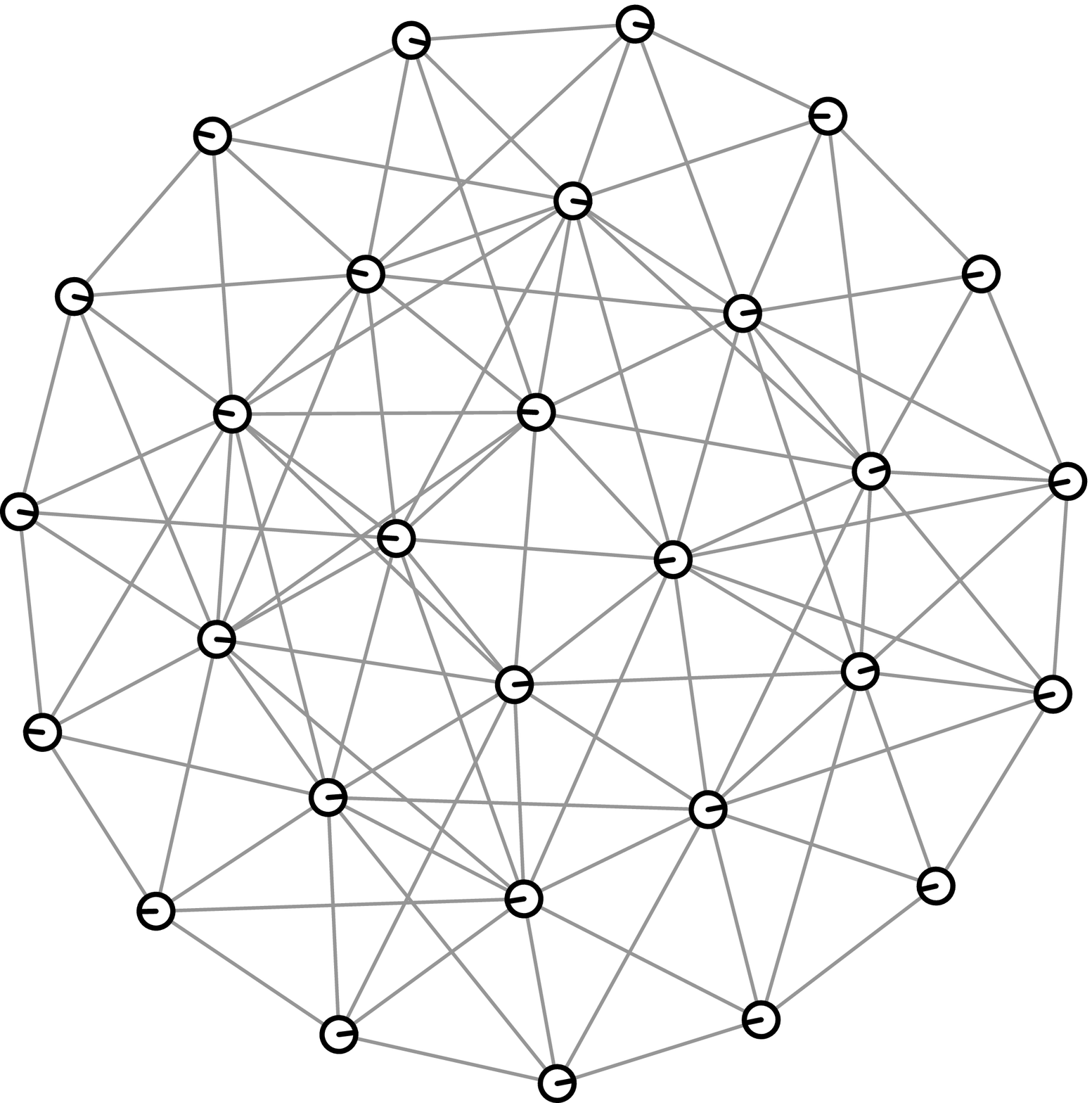}
	\caption{After}
\end{subfigure}
\caption{The base swarm forms the swarm similar to a water drop}
\label{fig:baseswarm}
\end{figure}

\begin{figure}[tbh]
	\centering
	\includegraphics[width=0.8\columnwidth]{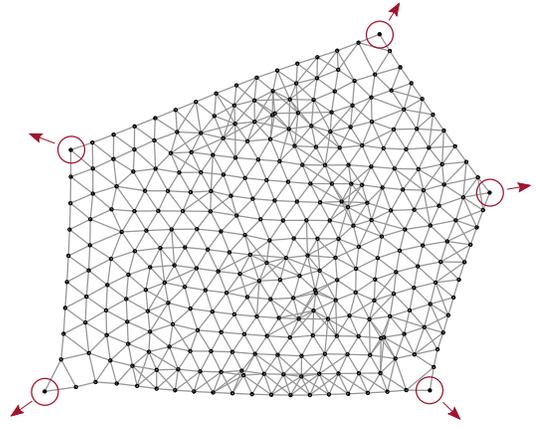}
	\caption{The base behavior without leader forces has trouble with staying connected after losing convexity.}
	\label{fig:baseleader}
\end{figure}

The base swarm is similar to a water droplet and converges towards a circle after some time.
The robots are well connected to the swarm and there are no attachments, as can be seen in Figure~\ref{fig:baseswarm}.
%
%
However, for diverging leaders the base behavior (movement consensus by flocking) 
without any other forces rapidly loses connectivity when the target density no longer suffices to cover the convex hull of leader robots. 
Figure~\ref{fig:baseleader} depicts a situation in which the swarm is about to lose convexity.    
For stronger control and more variable shapes, leader forces are introduced.

\subsection{Leader Forces}
\label{sec:alg:leader}
\begin{figure}[tbh] 
  \centering
  \includegraphics[width=0.45\textwidth]{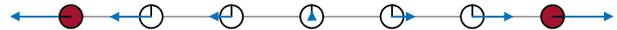}
  \caption{A one-dimensional scenario with two leaders (red) moving in opposite directions.}
  \label{fig:leadersmoothing}
\end{figure}

A single leader constitutes the simplest form of swarm control.
In this case the swarm motion is determined by the leader's velocity.
With multiple (possibly antagonistic) leaders, the swarm is not just steered, 
but may be stretched to the limit until connectivity is lost.
Therefore, each robot needs to find an appropriate balance between the influence of different leaders.
For $\ell\in\mathcal{L}$, let $c_\ell: \mathcal{R}\rightarrow\mathbb{R}^2$ be the force
on a specific robot and let $d_\ell:\mathcal{R}\rightarrow \mathbb{N}$ be its distance to $\ell$. 
The leader forces on robot $r$ are combined as follows:
\[ \sum_{\ell\in \leaderset}c_\ell(r)\frac{d_\ell(r)^{-1}}{\sum_{\ell'\in \leaderset}d_{\ell'}(r)^{-1}}.\]
See Figure~\ref{fig:leadersmoothing} for an illustration.

\begin{figure}[tbh] 
  \centering
  \includegraphics[width=0.4\textwidth]{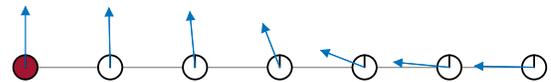}
  \caption{With increasing distance to the leader, the effect shifts from velocity matching to leader pursuit.}
  \label{fig:leaderforce}
\end{figure}

There are two ways of following a leader: either by matching its velocity or by moving towards it.
Velocity matching preserves the overall shape of the swarm, but fails with multiple leaders.
However, because the velocity information needs to be passed between robots with noisy sensors, there are accumulated losses in accuracy with each hop.
On the other hand, moving towards the leader causes a deformation of the swarm
and can be used to control its shape when multiple leaders are used, but
regions close to the leaders suffer from ``compression'', which can be harmful.  A
combination of both methods with a smooth transition between velocity matching
close to the leaders and leader pursuit when further away (see
Figure~\ref{fig:leaderforce}) has a positive influence in the context of multiple leaders,
both on accuracy and the overall swarm shape.

In order to achieve the combination of movement \emph{with} the leader and \emph{towards} the leader, three public variables are used for each leader.
The \textbf{leader distance} is the minimum hop count to the leader.
Let $\operatorname{pred}(r)$ be the predecessor in a minimum-hop tree to the leader, which can be the leader itself.
The \textbf{leader velocity} is the one of $\operatorname{pred}(r)$ for a non-leader, and the robot's own velocity for the leader.
The \textbf{leader direction} is a normalized direction vector calculated incrementally from the direction to $\operatorname{pred}(r)$ as follows:
Each robot takes the \emph{leader direction} of its $\operatorname{pred}(r)$ and merges it with the normalized direction to $\operatorname{pred}(r)$.
If $\operatorname{pred}(r)$ is the leader, only the normalized direction to it is used.
For computing the leader force, the \emph{leader direction} is scaled to the length of the \emph{leader velocity} and then combined with a \emph{leader distance}-sensitive weighting.

Additionally we provide leaders with too few neighbors with an attraction force, so they do not lose connection to the swarm.
This attraction spreads over some distance, but decreases exponentially.


\subsection{Stability Improvement}
\label{sec:alg:stability}
\begin{figure*}
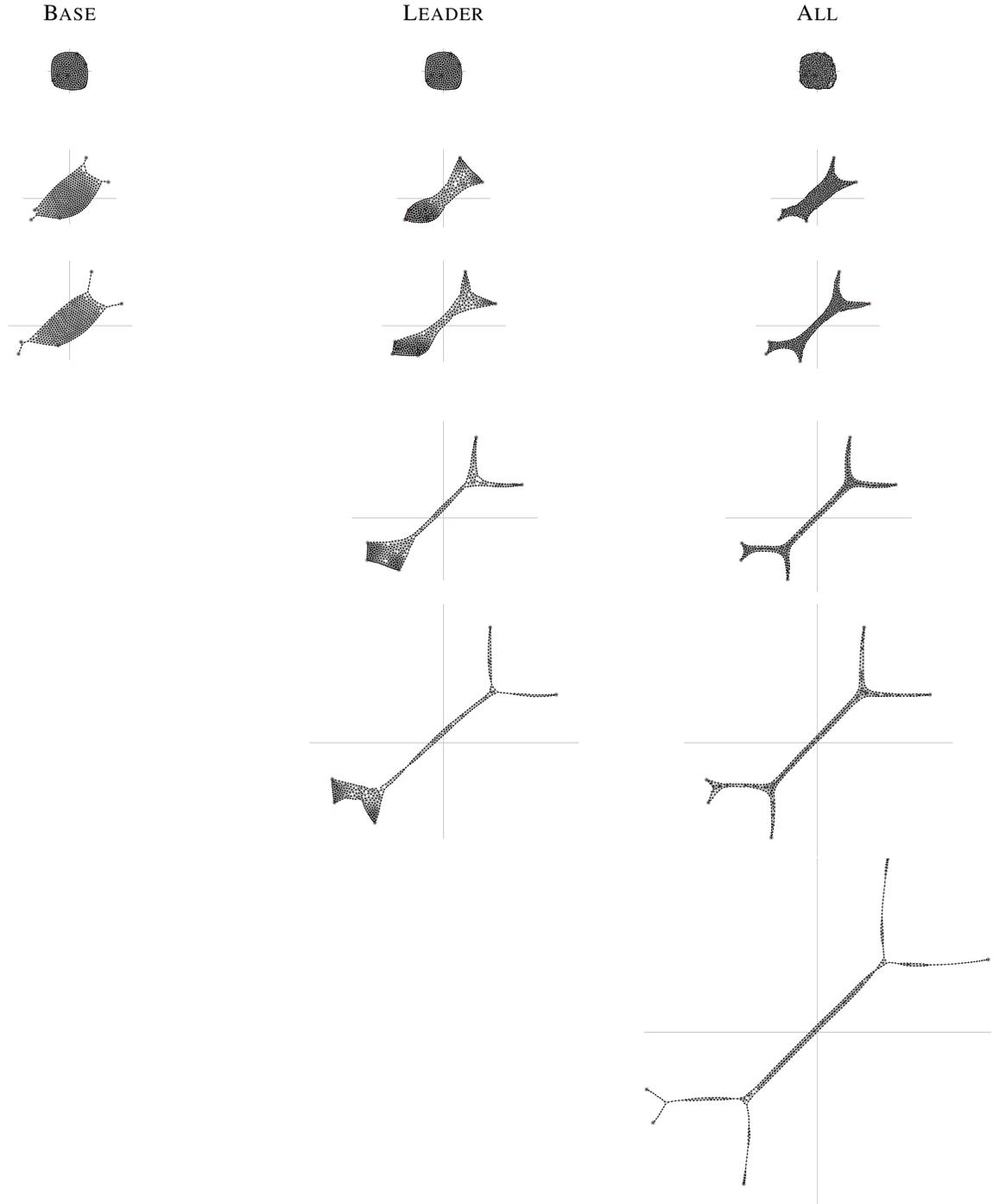
 
  \centering
  \newcommand{\swarm}[2]{\includegraphics[width=0.3\textwidth,trim=20mm 20mm 20mm 20mm,clip=true]{./failure_rate_0_#2_#1.pdf}}
  \newcommand{\timelabel}[1]{{}} 
\begin{tabular}{ccc}
\vspace*{-2cm}
 \sc Base & \sc Leader & \sc All\\
\vspace*{-3.5cm}
\swarm{200}{Base} & \swarm{200}{Leader} & \swarm{200}{All} \\
\vspace*{-3.5cm}
\swarm{2000}{Base} & \swarm{2000}{Leader} & \swarm{2000}{All} \\
\vspace*{-2.5cm}
 \swarm{3000}{Base} &  \swarm{3000}{Leader} & \swarm{3000}{All} \\
\vspace*{-2cm}
 & \swarm{5000}{Leader} & \swarm{5000}{All} \\
\vspace*{-1cm}
 & \swarm{7600}{Leader} & \swarm{7600}{All} \\
 & & \swarm{12000}{All}\\ 
\end{tabular} 
  \caption{A comparison of strategies for the same example, for a swarm with $n=400$ and failure rate $0$. As indicated, columns correspond to strategies {\sc Base}, {\sc Leader}, and {\sc All}.
Rows show the swarms at times $T=200$, $T=2000$, $T=3000$, $T=5000$, $T=7600$, $T=12,000$, with 60 steps per simulated second. When a swarm is no longer shown, it has become disconnected
right after the previous time step.}
  \label{fig:comparison}
\end{figure*}

Near Steiner points, connections along concave swarm boundaries may be stretched by boundary forces.
When the involved edges approach the upper bound for communication, connections may be disrupted, 
to the point where the swarm loses connectivity. 
By adding a thickness-dependent compression force, we reduce neighbor distances without 
influencing the Steiner-tree shape of the swarm;
in effect, this works similar to compression stockings.
In the following, we give a heuristic for thickness computation and compression. 
In order to let the flocking algorithm handle this compression without destroying the regular distribution, 
we sketch a density distribution heuristic later in this Section. 
A comparison of a swarm with and without the stability improvement can be seen in Figure~\ref{fig:comparison};
Figure~\ref{fig:comparison2} shows a comparison for the same scenario with failure rate $0.008$ per second and robot.

\begin{figure*} 
  \centering
  \newcommand{\swarme}[2]{\includegraphics[width=0.4\textwidth,trim=20mm 20mm 20mm 20mm,clip=true]{./failure_rate_13e-5_#2_#1.pdf}}
\begin{tabular}{cccc}
\vspace*{-1.5cm}
 \hspace*{-3cm}$T=4200$\hspace*{-3cm} & $T=4400$ \hspace*{-3cm}& $T=5400\hspace*{-3cm}$ & $T=5600$ \hspace*{-3cm}\\
\vspace*{-1.5cm}
 \hspace*{-3cm}\swarme{4200}{Leader} \hspace*{-3cm}& \swarme{final}{Leader} \hspace*{-3cm}& \hspace*{-3cm}& \hspace*{-3cm}\\
 \hspace*{-3cm}\swarme{4200}{All} \hspace*{-3cm}& \swarme{4400}{All} \hspace*{-3cm}& \swarme{5400}{All} \hspace*{-3cm}& \swarme{final}{All} \hspace*{-3cm}\\
\end{tabular} 
  \caption{A comparison of strategies for the example from Figure~\ref{fig:comparison}, for a swarm with $n=400$, with 60 steps per simulated second and failure rate $0.008$ per second. 
The upper line shows the swarm with strategy {\sc Leader}, the lower shows strategy {\sc All}. As shown, the swarm loses connectivity at $T=4400$ ({\sc Leader}),
or $T=5600$ ({\sc All}).}
  \label{fig:comparison2}
\end{figure*}

\paragraph{Thickness Contraction}
\label{sec:alg:stability:thickness}

\begin{figure}
	\centering
	\includegraphics[width=0.45\textwidth]{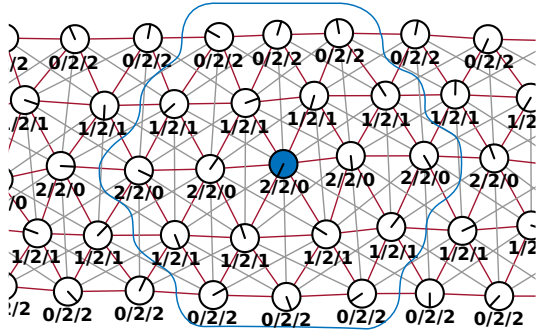}
	\caption{Thickness determination ($b(r)/t(r)/h(r)$) 
for a limb part. The red edges fulfill the Gabriel graph condition. A largest \emph{hop circle} is marked in blue.}
	\label{fig:thickness}
\end{figure}

We define the local thickness at a robot as the radius of the largest hop circle containing it.
A hop circle of radius~$h$ with robot~$c$ as circle center is the set of all robots with a hop count
$\leq h$ to $c$; only robots with distance equal to $h$ may be on the boundary.
An example is highlighted in blue in Figure~\ref{fig:thickness}.

The relationship between geometric thickness and boundary hop distance may be
distorted by long connections that skip over robots.  This can be avoided by
only considering edges that
fulfill the edge condition of the Gabriel graph, meaning that no robot is
allowed to be closer to the midpoint of an edge than the robots connected by it.
In principle, the resulting communication graph equals the Gabriel Unit Disk
Graph; this is the case when degenerate cases with line-of-sight obstructions are ignored.
We denote the corresponding reduced neighborhood of a robot~$r$ as $N'_r$.

The following method is a simplified implementation of the thickness metric above, which performed well enough in simulation.
It gets by with only three public variables; all circles with its center within a larger circle are ignored.

For this heuristic evaluation of the thickness~$t(r)$ of a robot~$r$, we need the hop distance~$b(r)$ from the boundary
and the circle center distance~$h(r)$.
Computing the hop distance to the boundary for each robot can easily be achieved
by setting $b(r)$ to 0 for all robots on the boundary, while all others take the minimum of their neighbors plus one, as follows
\[b(r)=\begin{cases} 0 & \text{$r$ on boundary}\\\min\{b(n)+1\mid n \in N'_r\} & \text{else}\end{cases}\]
Small holes, that occur frequently but also vanish quickly, are excluded from the boundary, otherwise the value can become too instable.
The thickness $t(r)$ is determined as the maximum $b(r)$ within some range $h(r)$, as follows.
\[t(r):=\max \{ \{b(r)\} \cup \{t(n)\mid n\in N'_r \wedge t(n)+\lambda \geq h(n)\}\},\]
where $\lambda\in \mathbb{N}$ is a small constant (e.g. $\lambda=2$) that tackles the problem of irregular boundaries.
If $r$ is a circle center ($t(r)=b(r)$), then the circle center distance~$h(r)$ is $0$. 
Otherwise, \[h(r):=\min\{ h(n)+1 \mid n\in N'_r\wedge t(n)=t(r) \}\]
An example is shown in Figure~\ref{fig:thickness}.


Based on this thickness $t(r)$, the described compression force grows linearly with this $t(r)$.
It acts only on robots of large boundaries, so that small holes are not prevented from closing.

\begin{figure}[tbh]
  \centering
  \includegraphics[width=0.7\columnwidth]{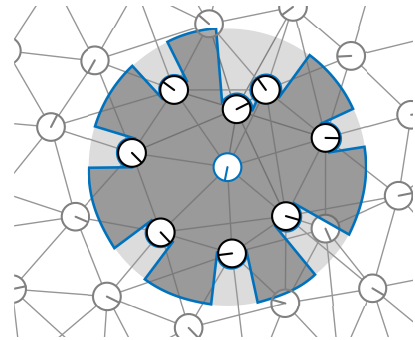}
  \caption{The observable area of a robot. The impact of hidden robots intersecting this area is ignored.}
  \label{fig:observablearea}
\end{figure}

\paragraph{Density}
\label{sec:alg:stability:density}
The local density of a robot refers to the number of neighbors in relation to its observable area as shown in Figure~\ref{fig:observablearea}.
By introducing an attraction to low and repulsion from high local density neighbors, the overall swarm density is maintained at a specific homogeneous level.

It is determined by dividing the number of neighbors by the roughly calculated observable area, cf.~Figure~\ref{fig:observablearea}. 
%
In order to avoid lumps, robots in collision range are weighted higher.
Dealing with the exterior area requires particular care, because its inclusion or exclusion from the calculation skews the results.
If the exterior area is included, boundary robots automatically get a lower density; if it is excluded, the density becomes too high.
We account for this by considering the exterior area of a robot as the area between the two adjacent boundary neighbors.
For overall balance, we assume its space to be the average space between two clockwise sequential neighbors that do not form an exterior area.
A robot can lie on multiple boundaries or multiple times on the same; however, this is a sign of a sparse distribution, so we only disregard the largest one.
All further exterior areas are fully included and thus lower the density.

The calculated observable area is sometimes not quite accurate, as the local knowledge is very limited.
Small heterogeneities can let the values vary strongly.
In order to improve the value, each robot first calculates its own value, but afterwards average this origin value with the origin values of the neighbors.
This averaged value is used to determine the attraction and repulsion forces.



Let $\rho(r')$ be the averaged local density of robot~$r'$, $\varrho$ the optimal density, and $N_r$ the neighbors of $r$. Then the density distribution force for a robot~$r$ is given by
\[ \sum_{n\in N_r} \overline{p}_r(n)*\phi(\rho(n) -\varrho),\]
where $\phi(x)=x^3/|x|$, and $\overline{p}_r: \mathcal{R}\rightarrow
\mathbb{R}^2$ is the direction from robot~$r$ to a neighbor with the length of
the distance for $\rho(n)\leq \varrho$; otherwise, it is of range minus distance. 
We do not apply this force to robots on the boundary.


\section{An Analytic Result}
\label{sec:theo}

Before describing the performance of our approach simulation results, we discuss
a related result from theoretical computer science, showing the analytic difficulty
of our underlying scenario, even for a centralized, static offline scenario without node failures.
In this setting, Efrat et al.~\cite{efg-iaarp-08} considered the {\em relay placement problem},
in which a given, static set of transmitters (called {\em terminals}) with limited communication
range must be connected by a set of more powerful {\em relays}; the objective is
to minimize the number of these relays for achieving connectivity. Clearly, this corresponds
directly to the achievable scaling factor for which a connected arrangment is possible:
The size of the arrangement is basically linear in the number of relays.

\begin{figure*}[th]
\vspace*{1cm}
        \centering
        \includegraphics[width=0.72\textwidth]{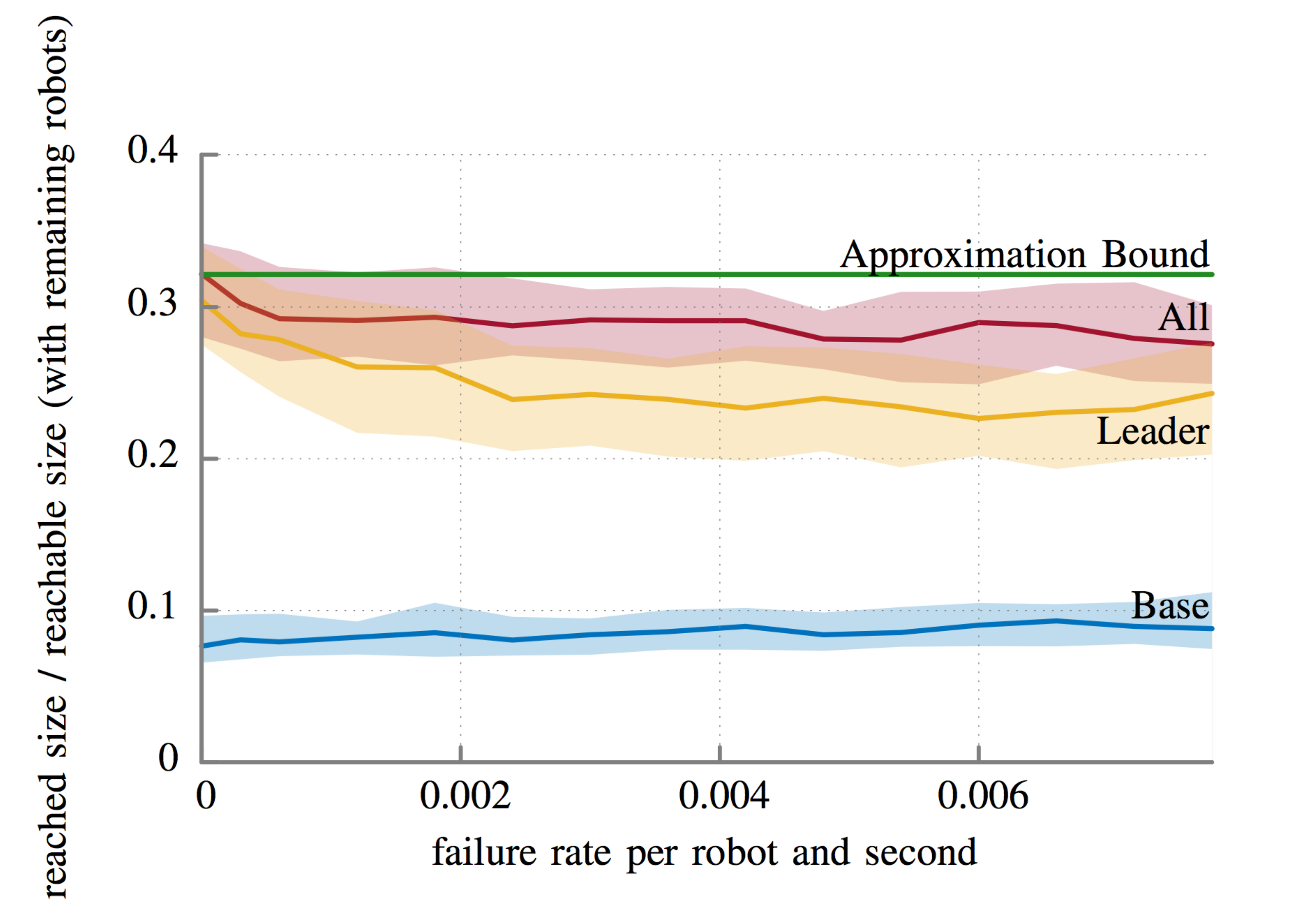}
\caption{Relative performance of the different strategy combinations,
measured by achievable Steiner tree size before disconnection occurs, compared
to a hypothetical static offline optimum for the remaining live robots. Shown are median
(bold) along with first and third quartiles. The failure rate is the
probability of {\em each} robot to die within the next simulated second, consisting of 60 time steps.
Clearly, the strategies are robust and adaptive; the full set
of strategies does particularly well in adjusting to leader motion and robot failures.}
\label{fig:performance}
\end{figure*}

As a generalization 
of the geometric Steiner tree problem, minimum relay placement is NP-hard.
To this date, the best known approximation factor for relay placement is the following.

\bigskip
{\bf Theorem~IV.1} (Efrat et al.~\cite{efg-iaarp-08})
There is a 3.11-approxi\-ma\-tion algorithm for minimum relay placement.

\bigskip
Note that this is a result for a guaranteed worst-case performance
of an algorithm, so we can hope to do better in specific settings.
However, we are also faced with a large number of additional difficulties that
make things much more difficult:
distributed setting, central control, dynamic movement of terminals
the necessity to make changes dynamically without losing connectivity, as well as
node failures.

\section{Simulation Results}
\label{sec:experiments}

We validated our approach by conducting experiments with a set of 
five leaders stretching out a swarm of 400 robots until it disconnects.
The performance is measured against the length of the minimal 
Steiner tree on disconnection (calculated by the Geosteiner 
software \cite{warme2001geosteiner}), divided 
by the theoretically maximal possible length estimated by 
$|\mathcal{R}'|*\operatorname{range}$, where $\mathcal{R}'$ 
are the robots that did not fail yet.
This would correspond to an optimal but extremely fragile 
Steiner tree in which {\em any} node failure disconnects the swarm. 
Thus, the best possible value of 1 is completely elusive,
in addition to being the result of an NP-hard offline optimization
problem.

For comparison we tested three configurations: 
\Base---only the base behavior as discussed in Section~\ref{sec:alg:base}; 
\Lead---the basic behavior enriched by leader forces as discussed in 
Section~\ref{sec:alg:leader};
\All---the final configuration that also incorporates Density and 
Thickness Contraction as presented in Section~\ref{sec:alg:stability}.

Our benchmark tests were carried out with 60 iterations per simulated 
second. We used parameters that correspond to those of the r-one robots
of Rice University~\cite{rone}:
robot diameter is $\SI{10}{\cm}$, communication range is $\SI{1.2}{\metre}$. 
The maximal robot velocity is $\SI{1}{\metre\per\second}$.
To account for the different source of leader motion, they were
limited to at most $\SI{0.25}{\metre\per\second}$,
giving the swarm robots the opportunity to react.


For each configuration we conducted 100 random trials on a range of different failure rates; note that 
a failure rate of $0.006$ per second corresponds to an expected lifetime of about 167 seconds, meaning that out of 400 robots,
on average about every 0.4 seconds one of them breaks down for good.
Figure~\ref{fig:performance} 
depicts the resulting performance for all three strategies; in each case, we show the median performance,
with corridors around the bold curves indicating first and third quartiles. The top part of Figure~\ref{fig:performance}
gives the performance relative to a hypothetical offline optimum {\em without} robot failures, which is extremely fragile:
as this solution is only a tree, {\em any} robot failure or uneven distribution will immediately disconnect it.
The ratio of 0.3215 (corresponding to the performance of a 3.11-approximation algorithm for relay placement)
is also indicated for better reference. The bottom part of Figure~\ref{fig:performance} gives the relative performance,
compared to a hypothetical optimum that can only use the remaining live robots.
It is clear to see that the strategies appear to be relatively robust against
sudden disconnection due to fatal robot failure events, indicating excellent
ability to adapt. 

Comparing the individual strategy components, the results show that leader forces already produce decent swarm behavior,
with survivability four times higher than for the base forces.
Without robot losses, it reaches about 30\% of the length of the hypothetical optimum, which is quite close to the theoretical
approximation ratio. With robot failures, the performance gets weaker with increasing failure probability.
The variant with additional stability improvement is slightly better without failures, but is clearly more robust against robot losses.



\section{Conclusion}
\label{sec:conclusion}

We have demonstrated how local methods for maintaining cohesiveness and connectivity
of a robot swarm can achieve remarkable results, even in the presence of exterior
forces and frequent, permanent robot failures. 

There are numerous possible and interesting extensions. 
One of them is to extend our methods to heterogeneous
swarms with different kinds of robots. In that setting, an even more structured, hierarchical
approach may be able to combine the strengths of centralized methods 
(which are better suited to keep track of unbalanced situations) with the benefits
of decentralized mechanisms (which are more robust against failure of key components).
Clearly, this looks promising in scenarios in inhomogeneous environments, in which
larger-scale, catastrophic events may cause rapid resource redistribution.
Other challenges include mastering more complex tasks, such as dealing
with obstacles, or performing collective transportation of objects by a swarm~\cite{RubensteinCWHMN13}.

\section*{Acknowledgment}
We thank James McLurkin and SeoungKyou Lee for many helpful conversations.

\addtolength{\textheight}{-12cm}   



%
%
%


\bibliographystyle{IEEEtran}
\bibliography{bibliography}

\end{document}